\documentclass[10pt,twocolumn,letterpaper]{article}

\usepackage{wacv}              % To produce the CAMERA-READY version
\usepackage{graphicx}
\usepackage{amsmath}
\usepackage{amssymb}
\usepackage{booktabs}

\usepackage{enumitem}
\usepackage[ruled,vlined,linesnumbered,noend]{algorithm2e}
\usepackage{multirow}
\usepackage{tabularx}
\usepackage{pifont}% http://ctan.org/pkg/pifont

\usepackage[noend]{algpseudocode}
\usepackage{etoolbox}
\makeatletter
\patchcmd{\@algocf@start}% <cmd>
  {-1.5em}% <search>
  {0pt}% <replace>
  {}{}% <success><failure>
\makeatother

\usepackage[pagebackref,breaklinks,colorlinks]{hyperref}
\newcommand{\cmark}{\ding{51}}%
\newcommand{\xmark}{\ding{55}}%

\usepackage[capitalize]{cleveref}
\crefname{section}{Sec.}{Secs.}
\Crefname{section}{Section}{Sections}
\Crefname{table}{Table}{Tables}
\crefname{table}{Tab.}{Tabs.}

\usepackage{xspace}
\newcommand{\Design}{\textit{$\mathsf{RS2G}$\xspace}}    
\def\wacvPaperID{1924} % *** Enter the WACV Paper ID here
\def\confName{WACV}
\def\confYear{2024}

\begin{document}

%%%%%%%%% TITLE - PLEASE UPDATE
\title{\vspace{-6mm}
RS2G: Data-Driven Scene-Graph Extraction and Embedding for Robust Autonomous Perception and Scenario Understanding\vspace{-5mm}}

\author{Junyao Wang, Arnav Vaibhav Malawade, Junhong Zhou, \\
Shih-Yuan Yu, Mohammad Abdullah Al Faruque\\
University of California, Irvine\\
Irvine, United States, 92697\\
{\tt\small \{junyaow4, malawada, junhonz2, shihyuay, alfaruqu\}@uci.edu}
% For a paper whose authors are all at the same institution,
% omit the following lines up until the closing ``}''.
% Additional authors and addresses can be added with ``\and'',
% just like the second author.
% To save space, use either the email address or home page, not both
% \and
% Arnav Vaibhav Malawade\\
% Institution2\\
% First line of institution2 address\\
% {\tt\small secondauthor@i2.org}
\vspace{-3mm}
}
\maketitle

%%%%%%%%% ABSTRACT
%\vspace{-2mm}
\begin{abstract}
\vspace{-2mm}
{Effectively capturing intricate interactions among road users is of critical importance to achieving safe navigation for autonomous vehicles.
While graph learning (GL) has emerged as a promising approach to tackle this challenge, 
existing GL models rely on predefined domain-specific graph extraction rules that often fail in real-world drastically changing scenarios.}
%often fail in real-world dynamically changing scenarios
% % Graph learning (GL) has been introduced as a promising approach to capturing intricate interactions among road users and achieving effective navigation for autonomous vehicles.
% However, existing GL models rely on predefined domain-specific graph extraction rules 
% %However, existing graph extraction relies on predefined domain-specific rules 
% that often do not capture the optimal graph representations.  
Additionally, these graph extraction rules severely impede the capability of existing GL methods to generalize knowledge across domains.
{To address this issue, we propose \textbf{RoadScene2Graph (RS2G)}, an innovative autonomous scenario understanding framework with a novel data-driven graph extraction and modeling approach that dynamically captures the diverse relations among road users.}
%To address this issue, we propose \textbf{RoadScene2Graph (RS2G)}, a novel data-driven graph extraction and modeling approach that dynamically identifies diverse relations among road users to enhance autonomous scenario understanding. 
Our evaluations demonstrate that on average RS2G outperforms the state-of-the-art (SOTA) rule-based graph extraction method by $4.47\%$ and the SOTA deep learning model by $22.19\%$ in subjective risk assessment.
More importantly, RS2G delivers notably better performance in transferring knowledge gained from simulation environments to unseen real-world scenarios. %, indicating notably enhanced generalization capability.

%RS2G exhibits considerably better performance for %generalization and 
%Sim2Real transfer learning %utilizing our novel Transformer relation extractor
%, indicating an enhanced ability to generalize %transfer 
%knowledge gained from simulation datasets to unseen real-world scenarios.
%\textcolor{red}{When testing in the 1043-carla \cite{yao2020dota}, we surpass the rule-based graph extraction method on the risk assessment task (from 95.86\% to 97.13\%) and Sim2Real transfer learning  (from \% to 66.29\% in 1043-carla to 620-dash).}

\end{abstract}%\vspace{-2mm}

% % % %%%%%%%%% BODY TEXT
\vspace{-3mm}
\section{Introduction}
\label{sec:intro}
Ensuring road safety to support various driving conditions has emerged as a fundamental research topic in the domain of \textit{autonomous vehicles} (AVs) ~\cite{zhang2022adversarial, shao2023safety}. 
As human drivers naturally reason about interactions between road users to effectively navigate their environments~\cite{strickland2018deep},
a number of innovative works modeling human driving experience to enhance the safety and robustness of AVs have been proposed~\cite{liang2023visual, zhu2023understanding, hu2023planning, codevilla2019exploring, choi2019gaussian}. However, it remains challenging to understand and model the diverse relations among driving agents to achieve adequate performance for autonomous risk assessment~\cite{kim2017end, huang2018apolloscape}. 
%and interact with diverse driving scenarios. 
Additionally, AVs are typically trained and tested with simulations and synthetic data, while real-world scenarios are often much more dynamic and unpredictable. 
This necessitates a learning approach that (\romannumeral 1) adeptly captures and models interactions between road users and environments, and (\romannumeral 2) effectively transfers knowledge from training domains to real-world settings. % is of critical need.
%This necessitates models that can effectively generalize knowledge gained from simulation environments to real-world driving conditions without substantial performance degradation.

\begin{figure}[!t]
%\vspace{-1mm}
\centering
\includegraphics[width=\linewidth]{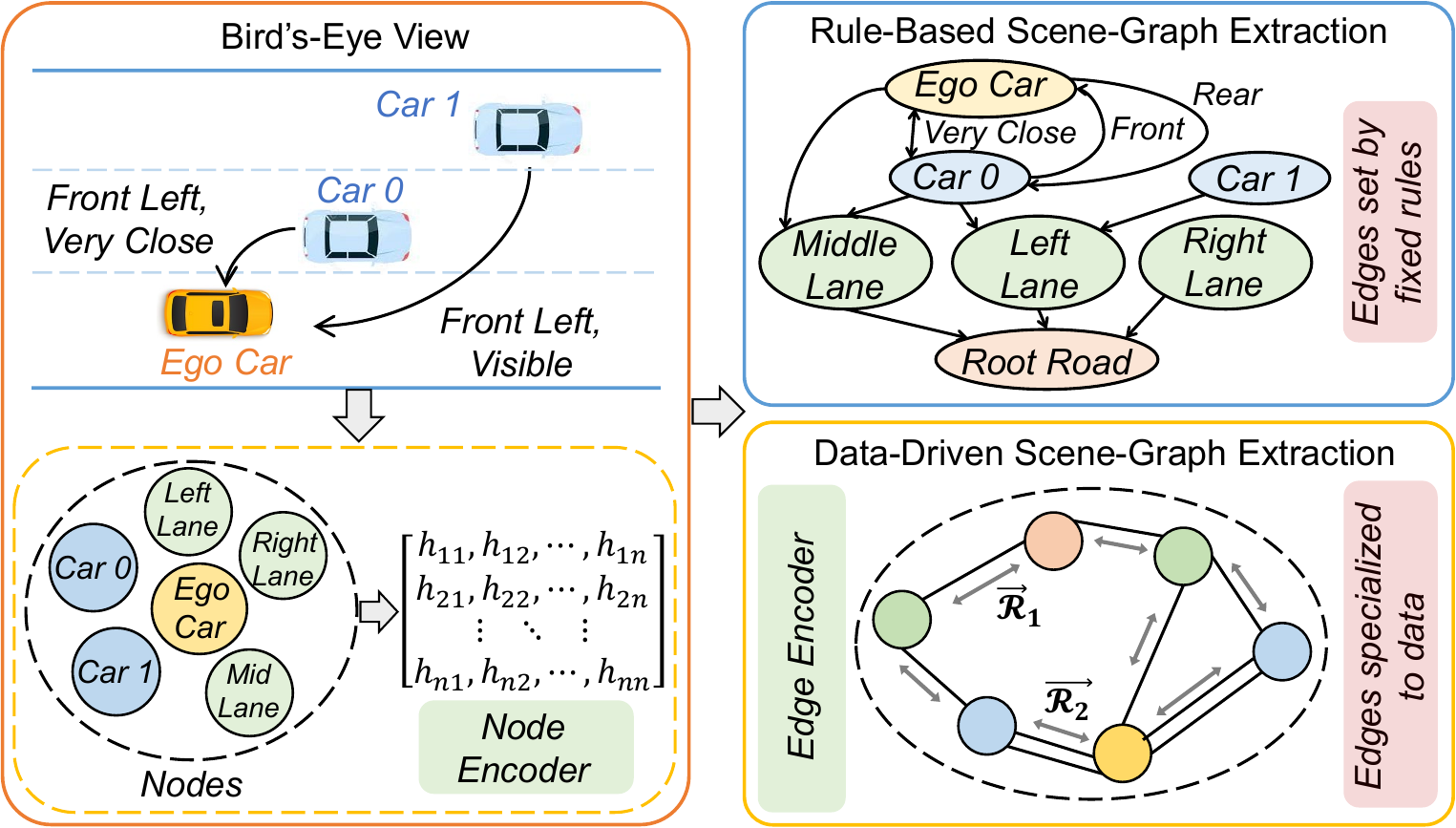}
\vspace{-6mm}
\caption{{Comparison of rule-based and data-driven scene-graph extraction. Both methods start with transforming objects to nodes with a node encoder, where each node has an attribute vector $\mathbf{h}$. %and generated its feature denoted as \textit{f}. 
The rule-based scene-graph extraction relies on fixed rules derived from expert knowledge; thus,
its generated edges typically have concrete physical meanings and the graphs are constrained by specific domains. In contrast, our data-driven scene-graph extraction represents diverse relations between nodes with vectors, e.g. $\mathcal R_1, \mathcal R_2$, which may contain more hidden information; therefore, its generated graphs are more dynamic and domain-adaptive. }}
%dynamically learns the optimal graph representation and hence generates more domain-adaptive graphs. In particular, our method represents diverse relations between nodes with vectors, e.g. $\mathcal R_1, \mathcal R_2$, which may contain more hidden information.}}
   \label{fig:motiv}
\vspace{-6mm}
\end{figure}

% \begin{figure*}[t]
%   \centering
%   {\includegraphics[width=0.9\textwidth]{./figs/motivation-crop}}
%   %\vspace{-2mm}
%   \caption{Comparing Rule-Based Scene-Graph Extraction and Data-Driven Scene-Graph Extraction}
% \vspace{-4mm}
%   \label{fig:motiv}
% \end{figure*}   

Many existing works leverage deep learning (DL) approaches, e.g., convolution neural networks (CNNs), to model human driving capabilities~\cite{djuric2020uncertainty, codevilla2019exploring, choi2019gaussian}. 
% and transfer knowledge gained from training domains to real-world driving scenarios
However, these methods often fail to account for information of high-level semantic scenes, thus performing less well in more complex or novel scenarios. 
Specifically, they rarely consider interactions between driving agents and environmental factors, e.g., effects of traffic signals on human behaviors~\cite{yu2021scene, djuric2020uncertainty, chen2017importance, blum2019fishyscapes}. 
Additionally, although existing DL-based data-driven approaches have demonstrated enhanced capabilities to generalize models across domains, i.e., improving the \textit{robustness} of models, these approaches often involve extraordinarily sophisticated models and massive labeled datasets~\cite{zhang2020causal, darestani2021measuring}. 
Thus, these approaches can be extremely expensive %as they incur significant model training and data storage costs, and can be less 
and hence impractical %in real-world settings 
as AVs are real-time systems with limited computational resources and onboard energy storage~\cite{li2021rts3d, choi2021shared}.
Besides, existing datasets are typically biased toward everyday driving situations but not diverse corner cases%involving higher risks
, often resulting in inadequate model performance in scenarios involving higher risks~\cite{codevilla2019exploring, huang2018apolloscape, zilberstein2015building}. 
%Therefore, a more resource-efficient learning method that can be well generalized to a variety of complex driving scenarios is of critical need.

%\textcolor{red}{In contrast to many existing solutions, our method not only demonstrates superior performance compared to traditional rule-based approaches on those rare cases like 620-dash dataset but also exhibits remarkable transferability. Specifically, when trained on standard everyday driving situations, like 271-carla and 1043-carla datset, our model consistently achieves the highest accuracy in handling rare scenarios, outperforming other methods that often degrade in such uncommon circumstances. }

%Graph learning (GL) has been introduced as a promising approach to capturing intricate relations among road users. 
In contrast to CNN-based models that directly extract visual features,  graph learning (GL) has emerged as a promising approach to explicitly capturing high-level interactions between visual features~\cite{casas2020spagnn, li2020learning, salzmann2020trajectron}.
{Prior works have shown that graph representations of driving scenarios extracted based on domain knowledge, referred to as rule-based~\textit{scene-graphs}, %as intermediate representations for GL 
enables effective modeling of diverse relations among road users and can potentially enhance autonomous scenario understanding~\cite{tian2020road, kochakarn2023explainable, mylavarapu2020understanding}. 
As demonstrated in Figure \ref{fig:motiv}, nodes of a rule-based scene-graph represent objects in a scene, e.g., lanes, vehicles, and traffic lights, while edges represent the types of relations, e.g., near and front left.
Additionally, it has been observed that sence-graphs can considerably improve data efficiency and transfer learning at AV safety-related tasks, e.g., collision prediction~\cite{malawade2022spatiotemporal, yu2021scene}.}
%Additionally, prior works have also demonstrated that graph representations of road scenes can considerably improve modeling capability, data efficiency, and transfer learning at AV safety-related tasks, e.g., collision prediction~\cite{ malawade2022spatiotemporal, yu2021scene}.
{Unfortunately, existing scene-graph extraction relies on predefined domain-specific rules,}
%GL-based models rely on predefined domain-specific graph extraction rules, % to %require expert knowledge 
%to define graphical structures, 
e.g., rule-based distance relations %, directional relations, 
and road topology~\cite{tian2020road, kochakarn2023explainable, yu2021scene, mylavarapu2020understanding}, resulting in rigid graphical structures. %, and rule-based directional relations in\cite{mylavarapu2020towards, yu2021scene}.
These graph extraction rules often fail to provide graph representations that are sufficiently expressive to achieve adequate performance.
%expressive graph representations of road scenes and fail to achieve adequate performance. 
The effectiveness of these rules also varies widely across domains, severely limiting their capability to generalize to %new domains or 
real-world scenarios absent from the training data.

To address this issue, we propose \textit{\textbf{RoadScene2Graph (RS2G)}}, an innovative autonomous scenario understanding framework with a novel data-driven graph extraction and modeling approach that \textit{dynamically} captures the diverse relations among road users. %\textit{\textbf{RoadScene2Graph (RS2G)}}, an innovative data-driven graph extraction and modeling approach that learns to \textit{dynamically} capture the optimal graph representations of a road scene.
As demonstrated in Figure~\ref{fig:motiv}, %for each image,
%rule-based graph extraction methods use fixed rules to predefine a set of graph edges and thus the extracted graphs are more static and \textit{domain-specific}. 
in contrast to rule-based graph extraction methods, %our data-driven graph extraction method, 
{$\Design$ represents graph edges with vectors that reflect the probability distribution of different object relations and capture the relations between nodes in a more granular manner.}
%\new{$\Design$ learns to specialize the set of graph edges with data-driven vectors, which may easily reflect more hidden relations between nodes and better represent the input data.}
The resulting graph representation hence becomes more expressive and \textit{domain-adaptive}. 
%to best represent the input data, thereby constructing more \textit{domain-adaptive} graph structures. 
%\new{Specifically, $\Design$ expresses relations by data-driven vectors that can easily reflect the hidden relationship.}
Specifically, we construct relations between nodes leveraging the Transformer as the edge encoder, since it has been proven that the attention mechanism of the Transformer is particularly powerful at capturing dependencies within inputs ~\cite{Vaswani2017AttentionIA}. 
We also utilize the variational autoencoder (VAE) to further improve the expressiveness of our graph representations and provide enhanced autonomous scenario understanding. 
%\todo{Additionally, we utilize Variational Autoencoder (VAE) to further generalize the representation of the edges to force the model to find more distinct and appropriate scene graph.}
% \textcolor{red}{In our approach, we leverage the encoder architecture of the Transformer model \cite{Vaswani2017AttentionIA} to serve as RS2G's edge encoder. The rationale behind this decision is grounded in the intrinsic capabilities of the Transformer. The Transformer's attention mechanism excels at capturing dependencies within an input sequence, making it particularly adept for our task. In the context of autonomous driving, understanding the spatial relationships between objects in a scene is vital for safe navigation. The relationships between, say, a pedestrian, a parked car, and an oncoming vehicle can provide crucial context for decision-making. Traditional methods, like the one from \cite{yu2021scene}, rely on fixed domain-knowledge rules, which can sometimes lack the flexibility to adapt to diverse and dynamic driving scenarios. }
The main contributions of our paper are listed as follows:
\vspace{-1ex}
%\todo{red}
\begin{itemize}[leftmargin = *]
%\setlength\itemsep{-1ex}
% \item \textcolor{red}{We introduce $\Design$, an innovative data-driven graph extraction and modeling approach that dynamically learns node embeddings and extracts the optimal graph representation of a road scene.}% to enhance scenario understanding of AVs.  
\item {We introduce $\Design$, an innovative autonomous driving risk assessment framework with a novel data-driven scene-graph extraction and modeling method. $\Design$ dynamically learns node embeddings and extracts the optimal graph representation of a road scene.}% to enhance scenario understanding of AVs.
\vspace{-1mm}
\item {To the best of our knowledge, $\Design$ is the first graph extraction method leveraging the powerful attention mechanism of the Transformer to capture relations and dependencies among road users. Our graph extraction technique  significantly enhances model performance for both subjective risk assessment and transfer learning from simulations to real-world scenarios.} 
\vspace{-1mm}
\item {We conduct detailed ablation studies regarding the benefits of each component of $\Design$ and further demonstrate the advantage of our data-driven graph extraction method %over the 
%{\color{cyan} previous approaches} 
in real-world autonomous scenario understanding.  
%% rule-based one.
}  
    %\item \textcolor{red}{We propose $\Design$, a data-driven graph extraction network based on GNN. It can well extract the features of the objects in the autonomous driving pictures and based on them, construct directed, heterogeneous multi-graphs with relations as edges to provide complete relation graphs for downstream tasks.}
    % \item We propose $\Design$, a data-driven graph extraction and learning methodology for autonomous driving. $\Design$ dynamically learns the structure of graphs and node embeddings, and hence effectively improves the capability to transfer knowledge from training domains to real-world scenarios.
    % \item \textcolor{red}{Our method introduce Transformer as edge encoder used to generate relationship between nodes which demonstrates great performance.
    % }
    % %\item Our proposed  $\Design$ outperforms the state-of-the-art DL-based and GL-based methods at subjective risk assessment.
    % \item We show that  $\Design$ can better transfer knowledge from simulation to real-world (e.g., Sim2Real) driving than state-of-the-art.
    % \item We illustrate how RS2G can model several rule-based relations simultaneously with each learned relation and how RS2G enables graph sparsity tuning with minimal performance impacts. 
\end{itemize}

\section{Related Works}
\label{sec:formatting}
\subsection{Interaction Modeling for Autonomous Driving}
Several recent works have demonstrated that explicitly modeling interactions between agents in dynamic environments can improve autonomous systems' capability to understand and reason about their environment~\cite{zilberstein2015building, langley2017explainable}. 
Multiple innovative learning frameworks using domain knowledge to extract graph representations of driving scenarios, i.e., \textit{scene-graphs}, for AVs have also been proposed. 
%have been proposed, focusing on rules for extracting graph representation of driving scenarios based on domain knowledge. 
%Specifically, in the AV use case, multiple works have proposed using domain-knowledge-derived rules to extract graph representations of driving scenarios, denoted as \textit{scene-graphs}. 
%These learning frameworks typically include (\romannumeral 1) a perception algorithm to identify the set of agents in the scene and their attributes, (\romannumeral 2) a set of graph extraction rules to build graph edges representing relations between nodes, and (\romannumeral 3) a DL-based graph model, such as the popular multi-relational graph convolution network (MR-GCN) to model heterogeneous multi-graphs.
In particular, \cite{tian2020road} proposes a rule-based graph extraction method encoding relationships %such as \textit{Same-lane}, \textit{Following}, \textit{Approaching}, \textit{Overtaking} 
between road users, and shows that this graph representation enables an effective autoencoder for inferring relationships between road users in unseen scenarios.
~\cite{kochakarn2023explainable} combines rule-based scene-graph extraction and MR-GCN to provide explainable predictions of future driver actions, and~\cite{yu2021scene} shows how a rule-based scene-graph improves risk assessment performance over CNN-based methods. Besides, \cite{mylavarapu2020understanding} uses a rule-based graph extracted from multi-modal sensor data to perform accurate driver action prediction.
Unfortunately, these methods rely on domain knowledge and rule-based graph extraction techniques, restricting them to their specialized tasks and data domains. 
In other words, different tasks require defining new rules, and each aforementioned work has a different set of rules. 
In contrast, our data-driven graph extraction approach eliminates such overhead, as we learn the graph extraction rules directly from the data, enabling high performance across tasks and data domains.

\subsection{Transfer Learning for Autonomous Driving}
Generalizing a trained model to unseen real-world scenarios without substantial performance degradation remains a critical challenge in autonomous driving. 
% Autonomous driving models are typically trained on a large synthetic dataset because developers can easily simulate many traffic conditions, road types, and driving scenarios. 
The term \textit{Sim2Real} describes the capability of a robotic system to %effectively 
transfer knowledge gained from simulation environments to real-world applications \cite{hofer2021sim2real}. 
Existing approaches addressing the transfer learning challenge can be mainly categorized as 
\textit{inductive} transfer learning and \textit{transductive} transfer learning \cite{niu2020decade}.
Inductive transfer learning involves learning a general set of rules from the source domain, e.g., training a supervised learning model, and applying them to the test domain. 
In contrast, transductive transfer learning utilizes some knowledge of the test domain to adapt a model trained in the source domain. 
Our work focuses on the inductive case as it better aligns with typical autonomous system applications, i.e., training machine learning (ML) models with processed or simulated data and testing them in diverse real-world settings. 
Several prior studies have leveraged inductive transfer learning to enhance model generalization capabilities.~\cite{zhang2021learning} transfers a CNN-based motion prediction model trained on pedestrian/vehicle trajectories to cyclist trajectories, and demonstrates that transferring knowledge from pedestrian motion prediction improves the performance of the cyclist motion prediction.~\cite{lamm2020vehicle} transfers knowledge from semi-supervised models with contrastive learning and teacher-student methods to improve trajectory prediction performance, and~\cite{maqueda2018event} evaluates transfer learning from traditional camera models to event camera models for steering angle prediction.~\cite{akhauri2021improving} transfers spatio-temporal features and uses salient data augmentation for better Sim2Real transfer performance in steering angle prediction and collision detection. More recently,~\cite{yu2021scene} demonstrates that graph-based scene modeling improves Sim2Real transfer performance compared to CNN-based methods, while its domain-specific graph extraction rules can limit the model's adaptivity across domains.

\section{Methodology}
%We elaborate the design of our proposed $\Design$ in this section. The architecture of $\Design$ is shown in Figure \ref{fig:archi}. 

%% In this section, we first introduce the problem formulation of subjective risk assessment, the task that necessitates robust scenario understanding and accurate modeling of interactions between agents. 
%formulate the problem we aim to solve with data-driven graph extraction and scenario modeling. We target subjective risk assessment as it is a task that necessitates robust scenario understanding and accurate modeling of interactions between agents. 
%Additionally, it is closely related to the important area of Advanced Driver Assistance Systems (ADAS) and can directly benefit tasks such as driver hand-off, collision avoidance, and emergency braking. 
%% We then explain the architecture of $\Design$ by elaborating our data-driven scene-graph extraction method and spatio-temporal graph modeling.
\begin{figure*}[t] 
%\vspace{-5mm}
\centering
    \includegraphics[width=\textwidth]{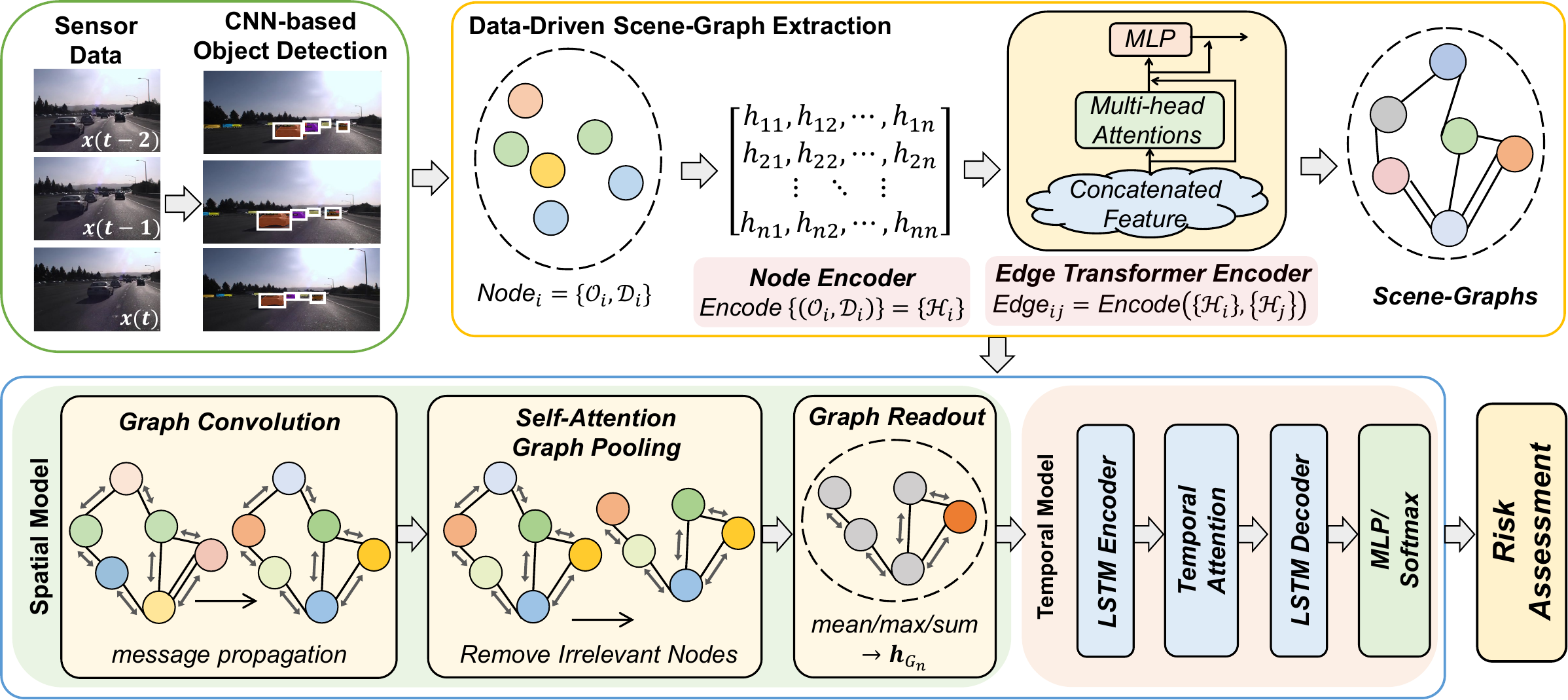}
    \vspace{-6mm}
    \caption{The Architecture of Our Proposed $\Design$. Given an input data sample, $\Design$ first extracts the set of objects and their attributes with a pre-trained CNN-based model. We then utilize our data-driven scene-graph extraction to generate a set of scene-graphs of the current scene and analyze it with our spatial-temporal embedding model. Finally, we utilize a multi-layer perceptron (MLP) to classify the risk of the driving scenario as risky or non-risky. }
    \label{fig:archi}
   \vspace{-2mm}
\end{figure*}

\subsection{Problem Formulation}
%abstract写上问题定··义，第三section更细致一点，不要太high-level
The problem of subjective risk assessment %can be modeled as inspired by~\cite{yu2021scene}. First, 
starts with a sequence of sensor data pre-processed by an object detection model and converted to a set of scene-graphs. These scene-graphs 
are then transformed into spatio-temporal embeddings for the subjective risk assessment. The overall system can be modeled as a binary classification task. 
Specifically, given that the input %to the model 
$\mathcal I$ is a sequence of sensor data, e.g., camera images, of length $\mathcal T$, a model $\Phi$ is constructed to generate the output $\mathcal Y$, % is used by the AV, 
i.e.,
\begin{equation}
    \mathcal Y=\Phi(\mathcal I); \quad \mathcal I = \{i_1, i_2, ..., i_{\mathcal T}\}
\end{equation}
\begin{equation*}
\begin{split}
% \mathcal Y=\Phi(\mathcal I); \quad \mathcal X = \{i_1, i_2, ..., i_{\mathcal T}\}\\
\textrm{and}%s.t. 
\quad \mathcal Y=\left \{
\begin{array}{ll}
     0, &  \text{if the driving sequence is safe}\\
     1, &  \text{if the driving sequence is risky}
\end{array}
\right.
\end{split}
\end{equation*}
where $\mathcal Y$ denotes the subjective risk indicator of the driving scene and  $\Phi$ represents the function mapping inputs $\mathcal I$ to $\mathcal Y$.
An ML model is often utilized to approximate the function defined by $\Phi$, where the inference output is denoted as $\hat{\mathcal Y}$.

Some ML methods, namely CNNs, directly operate on sensor inputs to produce output classifications $\hat{\mathcal Y}$. However, these approaches only model pixel-level features without considering inter-object relations for high-level objectives. 
%as discussed in Section \ref{sec:intro}. 
On the other hand, CNN-based models typically perform well for low-level tasks, such as object detection, since these tasks are less dependent on inter-object semantic relations. 
Therefore, for each input data sample $i_t$ in the sequence, i.e., $i_t\in \mathcal I$, we first utilize a pre-trained CNN-based object detection model $\Omega$ to efficiently extract the set of objects $\mathcal O_t$ and their attributes $\mathcal D_t$. 
We then utilize our graph extraction model $\Psi$ (elaborated in section \ref{subsec:graph_ext}) %these outputs are passed to our graph extraction model $\Psi$ 
to generate a scene-graph $\mathcal G_t$, i.e.,
% \begin{align}
%     & \mathcal O_t, \mathcal D_t = \Omega(i_t) \\
%     & \mathcal G_t = \Psi(\mathcal O_t, \mathcal D_t)
% \end{align}
\begin{equation}
    \mathcal O_t, \mathcal D_t = \Omega(i_t) \textrm{\space\space and \space\space}
    \mathcal G_t = \Psi(\mathcal O_t, \mathcal D_t)
\end{equation}
%Using pretrained CNN-based models $\Omega$ to extract objects from the scene enables us to efficiently produce scene-graphs from sensor data since we can focus our graph extraction model $\Psi$ on semantic relationship modeling between the extracted objects instead of low-level perception. 
We denote a \textit{scene-graph} as $\mathcal G_t = (\mathcal O_t, \mathcal A_t)$ and model it as a directed heterogeneous multi-graph since multiple types of edges can exist between any two nodes. 
$\mathcal O_t$ denotes the set of nodes and represents objects in a scene. The edges of $\mathcal G_t$ are represented by the adjacency matrix $\mathcal A_t$, and each value in $\mathcal A_t$ represents the type of the corresponding edge in $\mathcal G_t$. %, and the edges of $\mathcal G_t$ are represented by the adjacency matrix $\mathcal A_t$. In particular,  each value in $\mathcal A_t$ represents the type of the corresponding edge in $\mathcal G_t$, and multiple types of edges can exist between any two nodes. 
Once all the scene-graphs are extracted for the current scene, we analyze the collection of graphs $\mathcal G$ with our spatio-temporal graph embedding model ${\Phi}$ (elaborated in section \ref{subsec:graph_model}) to classify ${\mathcal Y}$ by its risk. 
Thus, the complete system can be modeled as
\begin{equation}
    \hat{\mathcal Y} = {\Phi}(\mathcal G) \quad
\end{equation}
where $\mathcal G = \{\Psi(\Omega(i_t)) \; \forall t \in \{1, 2, ..., \mathcal T\}\}.$
% \begin{equation}
%     \hat{\mathcal Y} = \hat{\Phi}(G) \quad \textrm{such that} \quad \mathcal G = \{\Psi(\Omega(i_t)) \; \forall t \in \{1, 2, ..., \mathcal T\}\}
% \end{equation}
% %pass the collection of graphs $G$ to our spatio-temporal graph embedding model $\hat{\phi}$ to make a risk classification $\hat{Y}$ for the driving scene. Thus, our complete system model can be represented as follows.
% \begin{equation}
%     \hat{Y} = \hat{\phi}(G) \quad s.t. \quad G = \{\Psi(\Omega(i_t)) \; \forall t \in \{1, 2, ..., T\}\}
% \end{equation}
% We elaborate on the implementation of $\Psi$, and $\hat{\phi}$ in Sections \ref{subsec:graph_ext} and \ref{subsec:graph_model}, respectively.

The architecture of our proposed $\Design$ is demonstrated in Figure \ref{fig:archi}. We elaborate on each component of $\Design$ in the rest of this section.

\subsection{Data-Driven Scene-Graph Extraction}%\vspace{-2mm}}
\label{subsec:graph_ext}
%3.2 intuitition重点，hyperparamter，data-driven的关系

% \begin{algorithm}[ht]
% \SetAlgoLined
% \LinesNumbered
% \DontPrintSemicolon
% \textbf{Input:} Objects $\mathcal O_t$ and their attributes $\mathcal D_t$ at time $t$.\\
% \textbf{Output:} Scene-graph $\mathcal G_t$ at time $t$.\;
% \SetKwFunction{Fgraphext}{$\Psi$}
% \SetKwProg{Fn}{def}{:}{}

% \Fn{\Fgraphext{$\mathcal O_t, \mathcal D_t$}}{
% $\mathcal H_t \gets \{ \}$\\
% $\mathcal A_t \gets \{\{\{ \}\}\}$ \Comment{initialize the adjacency matrix}\\
% \For{$o_j, \mathbf d_j \in \mathcal O_t, \mathcal D_t$}{
% $\mathbf h_j \gets Encode_{node}(o_j, \mathbf d_j)$\Comment{node encoding}
% $\mathbf H_t.append(\mathbf h_j)$
% }

% $\mathcal C \gets \mathcal H_t \times \mathcal H_t$\Comment{get all pair of nodes}\\
% \For{$r \in \mathcal R$}{
% \For{$\mathbf h_j, \mathbf h_k \in \mathcal C$}{
% \tcc{\texttt{get edges for relation }$r$}
% $\tau \gets Encode_{edge}(r, \mathbf h_j, \mathbf h_k)$ %\Comment{$n$ denotes the number of edge types}\;\\
% % n is the hpyer-parameter of transformer input
% $(\mathcal A_t)_{r,j,k} \gets \textit{MLP}(\tau)$\\
% %$(\mathcal A_t) _{r,j,k} \gets Encode_{edge}(r, \mathbf h_j, \mathbf h_k)$\;
% }
% }
% $\mathcal G_t \gets \{ \mathcal H_t, \mathcal A_t\}$\;%\tcp*{add nodes \& edges to G}
% \KwRet{$G$}\;
% }
% \caption{Data-Driven Scene-Graph Extraction}
% \label{alg:extractgraph}
% \end{algorithm}
% \setlength{\textfloatsep}{0pt}

%% new edits

Our methodology for extracting a scene-graph $\mathcal G_t$ from a set of objects $\mathcal O_t$ and their attributes $\mathcal D_t$ at a given time $t$ is demonstrated in Algorithm \ref{alg:extractgraph}. 
%We model our scene-graphs as directed, heterogeneous multi-graphs, where objects in the scene are represented by nodes and relationships between objects are modeled as edges.
%Each node contains an attribute vector and each edge is a directed edge with binary types, and multiple types of edges can exist between any two nodes. 
In contrast to the SOTA graph extraction method~\cite{yu2021scene} using fixed rules derived from domain knowledge to %define the conditions for constructing
construct graph edges, e.g., threshold-based distance relations and compass-based directional relations, {we propose an innovative %$\Design$ utilizes a 
data-driven edge encoder to generate \textit{domain-specialized} edge types. 
%\new{Our edge encoder \textit{dynamically} learns the rules for building edges between each pair of nodes, and thereby not only performs well in the training domain but also effectively generalizes to other domains.}
{Specifically, our approach starts with a node encoder model $Encode_{node}$ that converts the attributes $\mathbf{d}_j (\mathbf{d}_j\in\mathcal D_t) $ of each object $o_j (o_j\in\mathcal O_t)$ into a set of encoded node features $\mathbf{h}_j\in \mathcal H_t$ by its entity type, e.g., car and lane,  and coordinate, i.e. location, where $\mathcal H_t$ denotes the collection of encoded node features of all the objects at time $t$.}
We then concatenate feature vectors of each pair of nodes and send the resulting vector to an edge encoder, denote as $Encode_{edge}$ in Algorithm \ref{alg:extractgraph}, to infer if there is an edge of a given relation type $r\in \mathcal R$ %\textcolor{red}{,where R are the relationships learned by the model and r is decided by} node $j$ and node $k$
, given their features $\mathbf h_j$ and $\mathbf h_k$. {Here $\mathcal R$ denotes the set of all the possible relations between nodes, and $(j,k)$ represents all the possible combinations of two nodes in $\mathcal H$, i.e., $\{(j, k) | \forall j, k \in \mathcal H, j\neq k\}$.
Each relation type has a different set of learnable weights, and our edge encoder is responsible for computing these weights to find different rules for constructing each relation type. }
%\textcolor{red}{It can be explained that each relation type has a different set of learnable weights. Our edge encoder is responsible to find them, have the ability to learn different rules for constructing each relation type between nodes.}
{Intuitively, 
with a strong ability to capture complex non-linear mappings and hierarchical feature interactions, a multi-layer perceptron (MLP) can effectively discern relations between nodes and thus serve as a proper edge encoder.} %, denoted as $Encode_{edge}$ in Algorithm \ref{alg:extractgraph}.} %Besides, we also try Transformer as edge encoder which shows great performance not only in subjective risk assessment but also in transfer tasks.}%serve as a proper edge encoder as it can discern relationships between nodes due to its ability to capture complex non-linear mappings and hierarchical feature interactions.} %Moreover, we innovatively employ the Transformer as an edge encoder. This choice has proven effective, particularly in subjective risk assessment and transfer tasks, which are elaborated in \ref{subsec:risk_assessment} and \ref{subsec:transfer}.} 
\vspace{-1mm}
\begin{algorithm}[ht]
\SetAlgoLined
\LinesNumbered
\DontPrintSemicolon
\textbf{Input:} Objects $\mathcal O_t$ and their attributes $\mathcal D_t$ at time $t$.\\
\textbf{Output:} Scene-graph $\mathcal G_t$ at time $t$.\;
\SetKwFunction{Fgraphext}{$\Psi$}
\SetKwProg{Fn}{def}{:}{}

\Fn{\Fgraphext{$\mathcal O_t, \mathcal D_t$}}{
$\mathcal H_t \gets \emptyset$, $\mathcal A_t \gets \mathbf{0}_{n \times n}$
\Comment{initialize outputs}\\
\For{$o_j, \mathbf d_j \in \mathcal O_t, \mathcal D_t$}{
$\mathbf h_j \gets Encode_{node}(o_j, \mathbf d_j)$\Comment{node encoding}\\
$\mathcal H_t.append(\mathbf h_j)$
}

$\mathcal C \gets \mathcal H_t \times \mathcal H_t$\Comment{get all pair of nodes}\\
\For{relation $r \in \mathcal R$}  { %% \Comment{iterate on relation $r$}
\For{edge $(\mathbf h_j, \mathbf h_k) \in \mathcal C$}{
%% \tcc{\texttt{get edges for relation }$r$}
% $\tau \gets Encode_{edge}(r, \mathbf h_j, \mathbf h_k)$ %\Comment{$n$ denotes the number of edge types}\;\\
% n is the hpyer-parameter of transformer input

% $(\mathcal A_t)_{r,j,k} \gets \textit{MLP}(\tau)$

$(\mathcal A_t)_{r,j,k} \gets \textit{MLP}(Encode_{edge}(r, \mathbf h_j, \mathbf h_k))$

%$(\mathcal A_t) _{r,j,k} \gets Encode_{edge}(r, \mathbf h_j, \mathbf h_k)$\;
}
}
$\mathcal G_t \gets \{ \mathcal H_t, \mathcal A_t\}$\;%\tcp*{add nodes \& edges to G}
\KwRet{$\mathcal G_t$}\;
}
\caption{Data-Driven Scene-Graph Extraction}
\label{alg:extractgraph}
\end{algorithm}
\setlength{\textfloatsep}{8pt}

%\vspace{8mm}
\subsubsection{{Edge Encoder Inspired by the Transformer}}
%\vspace{-2mm}
To further enhance $\Design$'s capability to model complex relations among road users, 
we propose an innovative edge encoder based on the Transformer %, denoted as ${TransEncode}_{{edge}}$ in Algorithm \ref{alg:extractgraph}, 
to process the concatenated features. 
Transformer is an attention-based model that performs particularly well in capturing dependencies among input vectors, and naturally fits this edge prediction task that aims to model relationships between nodes. 
{Specifically, we first calculate the transformed information $\mathcal Q$ (query), $\mathcal K$ (key), and $\mathcal V$ (value) representations from node features as $\mathbf h_t \mathcal W_i^{\mathcal Q}, \mathbf h_t \mathcal W_i^{\mathcal K}, \mathbf h_t \mathcal W_i^{\mathcal V}$, where $\mathbf h_t$ represents the feature vector of a node at time $t$ and $\mathcal W_i^{\mathcal Q}$, $\mathcal W_i^{\mathcal K}$, $\mathcal W_i^{\mathcal V}$ are trainable parameters for $\mathcal Q, \mathcal K, \mathcal V$, respectively.}
We then utilize an attention mechanism to enable each node to understand its context and implicitly establish relationships with other nodes in the scene. 
We obtain the relation representation $\tau$ by:
\begin{equation}
        \tau = \mathcal Q \mathcal K^{\mathcal T}\mathcal V
\end{equation}
    %{where $\mathcal Q,\mathcal K, \mathcal V$ are the transformed information from the raw input calculated by {$XW_i^Q,XW_i^K,XW_i^V$} which contains potential hidden information of the nodes.}
Finally, we apply a multi-layer perceptron (MLP) to transform $\tau$ from dimension $\mathbf h_j$ to the total number of potential relations $|\mathcal R|$. % to accommodate the potential relations. 
The output of our data-driven scene-graph extraction is an $n \times n \times |\mathcal R|$ adjacency matrix $\mathcal A_t$, where $n$ denotes the total number of nodes and $|\mathcal R|$ denote the number of relation types.
Along with the node features $\mathcal H_t$, this adjacency matrix $\mathcal A_t$ forms the backbone of the scene-graph $\mathcal G_t$, offering a more dynamic representation of scenes.
% \textcolor{red}
{%By processing through $k\times v$ (where $v$ represents each node's value), we obtain a representation, denoted as \(T\), that captures these relationships. Subsequently, a multi-layer perceptron (MLP) is used to transform \(T\) from dimension \(h_j\) (denoted as \(n\)) to accommodate the dimension of potential relations \(R\). The outcome is an \(n \times n \times R\) adjacency matrix for \(n\) total nodes and \(R\) relation types. Combined with the node features \(H\), this matrix forms the backbone of the scene-graph \(G\), offering a more dynamic and data-driven representation of scenes crucial for advanced autonomous driving systems.}

\vspace{-3mm}
\subsubsection{Scene-Graph Generalization}
\label{subsubsec:graph_gene}
{In the context of autonomous driving, we aim to enable our model to comprehensively capture all the essential relations among road users, while avoiding excessively including minor details; otherwise, it may cause overfitting and impede the generalization capability of our model.}
Given the powerful nature of the Transformer model in capturing intricate dependencies and relations among inputs, instead of directly using the adjacency matrix $\mathcal A_t$ as our graph representations, we introduce the variational autoencoder (VAE) for regularization to enhance our model's capability to generalize across various driving conditions. Specifically, we reparameterizing each relation vector that represents edge types in $\mathcal A_t$ as 
%we introduce the variational autoencoder (VAE) to avoid overfitting and thereby enhance model's capability to generalize across diverse driving conditions. 
%In particular, instead of directly using the adjacency matric $\mathcal A_t$ as our graph representations, 
%we regularize $\mathcal A_t$ by reparameterizing each relation vector that represents edge types as 
%we regularize graph representations by reparameterize each edge vector of $\mathcal A_t$ as 
\begin{equation}
    z = \mu + \log(\sigma^2) \times \epsilon
\end{equation}
{where $z$ represents a new relation vector transformed from regularizing each relation vector in $\mathcal A_t$;}   $\mu$ and $\sigma^2$ indicate the mean and variance of each relation vector in $\mathcal A_t$, respectively, and $\epsilon$ represents random samples from a standard normal distribution, i.e., $\epsilon \sim \mathcal N(0,1)$. 
This serves as a form of ``bottleneck'' that enforces the model to capture the most salient features of the adjacency matrix.
Specifically, this reparameterization technique enables a back-propagation by incorporating more randomness into the model, hence ensuring a smooth gradient landscape for optimization. 
{Moreover, we involve the Kullback-Leibler (KL) divergence as a loss term to train our model. It not only ensures a more meaningful structure of the latent space to provide a more generalized, effective, and independent relation representation.}
%latent space has a meaningful structure, 
but also enables the model to generate more expressive adjacency matrices to enhance graph representations of road scenes.   
\subsection{Spatial-Temporal Graph Embedding Model}
\label{subsec:graph_model}
As demonstrated in Figure \ref{fig:archi}, our spatial-temporal model consists of three major components, a spatial model, a temporal model, and a risk inference component. 
The spatial model outputs the embeddings $\mathbf h_{\mathcal G_t}$ for each scene-graph $\mathcal G_t$, and the temporal model processes all input scene-graph embeddings, i.e., $\{\mathbf h_{\mathcal G_1}, \mathbf h_{\mathcal G_2}, \ldots,\mathbf h_{\mathcal G_{\mathcal T}}\}$, and produces the spatial-temporal embedding $\mathcal Z$. 
Then the risk inference components outputs each driving clip's risk assessment $\hat{\mathcal Y}$. 
% We model a series of scene-graphs for risk assessment by combining graph modeling capturing spatial information and sequence modeling capturing temporal information, as demonstrated in Figure \ref{fig:archi}. 
% Specifically, after extracting all the scene-graphs with Algorithm \ref{alg:extractgraph}, we utilize a spatial graph model to generate a set of graph embeddings. 
% % First, an object detection model extracts the set of objects in the scene, which are then processed into a graph using Algorithm \ref{alg:extractgraph}. The resulting graphs are passed to our spatial graph model to produce a set of graph embeddings. 
% These graph embeddings are then temporally modeled to produce a final risk assessment for the driving scene.% We elaborate on these components in the following subsections. %Our complete risk-assessment workflow is detailed in Section \ref{subsec:risk_assessment}.
\subsubsection{Spatial Graph Modeling}
We utilize a multi-relational graph convolutional network (MR-GCN) \cite{mylavarapu2020towards} to compute the embeddings and capture multiple types of relations on each scene-graph.
%Specifically, e
Specifically, in the \textit{message propagation} phase, each MR-GCN layer performs spatial graph convolutions \cite{kipf2016semi} on each graph $\mathcal G_t = \{\mathcal H_t, \mathcal A_t\}$ for all $t\in \{1, 2, \ldots, \mathcal T\}$ across a set of relation types $\mathcal R$, where $\mathcal T$ denote the length of the data sample. For each node $v \in\mathcal G_t$, the $l$-th MR-GCN layer updates the node embedding, denoted as 
$\mathbf{h}^{(l)}_v$, as
\begin{equation}
    \mathbf{h}^{(l)}_v = \mathbf{\Phi}_{\textrm{0}} \cdot
        \mathbf{h}^{(l-1)}_v + \sum_{r \in \mathcal R} \sum_{u \in \mathbf{N}_r(v)}\frac{1}{|\mathbf{N}_r(v)|} \mathbf{\Phi }_r \cdot \mathbf{h}^{(l-1)}_u,
\end{equation}
where $\mathbf{N}_r(v)$ denotes the set of neighbor indices to $v$ with relation $r$ in the adjacency matrix $\mathcal A_t$,  
$\mathbf{\Phi }_r$ reprsents the set of trainable weights for relation $r$ in MR-GCN layer $l$. 
{Since the $(l-1)$-th layer can directly influence the node representations in the $l$-th layer, MR-GCN applies another trainable transformation $\Phi_0$ to account for the self-connection of each node using a special relation~\cite{schlichtkrull2018modeling}. 
We initialize each node embedding $\mathbf h_v^{(0)}, \forall v\in \mathcal O_t$ by directly converting the node's type information to its corresponding one-hot vector. }
% and %
% % Since the information in $(l-1)$-th layer influences the node representation at $l$-th layer, the MR-GCN uses another trainable transformation $\mathbf{\Phi }_0$ to account for the self-connection of each node using a special relation~\cite{schlichtkrull2018modeling}.
% {$\mathbf{\Phi }_0$ is used to assign the self-connection of each node $\mathbf{h}^{(l-1)}_v$ itself using \textcolor{red}{a special relation}~\cite{schlichtkrull2018modeling}. }

Node embeddings typically become more refined and abstract as the number of MR-GCN layers increases, while the output of the features from earlier MR-GCN layers can be better generalized across domains~\cite{xu2018powerful}.{
Therefore, we utilize the node embeddings generated from all the MR-GCN layers. 
Specifically, we calculate the embedding of node $v$ at the final layer, denoted as $\mathcal{H}^{\mathcal L}_v$, by concatenating the features generated from all the MR-GCN layers, i.e., %such that the embedding of node $v$ at the final layer, denoted as $\mathcal{H}^{\mathcal L}_v$, is the concatenation of the features generated from all the MR-GCN layers, i.e.,},
\begin{equation}
\mathcal{H}^{\mathcal L}_v = \textbf{CONCAT}(\{\mathbf{h}^{(l)}_v\}|l=0, 1, ..., \mathcal L).
\end{equation}
where $\mathcal L$ denotes total number of layers. We denote the collection of node embeddings of \textit{scene-graph} $\mathcal G_t$ after passing through $\mathcal L$ layers of MR-GCN as $\mathcal X^{prop}_{t}$.

%Since our task is graph sequence classification, we 
We then utilize a graph pooling and readout operation to condense the set of node embeddings $\mathcal X^{prop}_{t}$ to a single, unified graph embedding $\mathbf h_{\mathcal G_t}$. Here we employ the \textit{self-attention graph pooling} operation~\cite{lee2019self}. Specifically, in the pooling layer, nodes are pooled according to the scores predicted from a trainable graph convolutional networks (GCN) layer, denoted as \textbf{SCORE}, as 
\begin{equation}
\mathbf{\alpha} = \textbf{SCORE}(\mathcal X^{prop}_{t}, {\mathcal A_{t}}) \quad\textrm{and}\quad \mathcal P = \mathrm{top}_k(\mathbf{\alpha}),
\end{equation}
% \begin{equation}
% \label{eq:pool}
% \mathcal P = \mathrm{top}_k(\mathbf{\alpha}), 
% \end{equation}
where $\mathbf{\alpha}$ represents the attention coefficient output by the graph pooling layer for each node in $\mathcal G_t$, $\mathcal P$ represents the top $k$ proportion of nodes ranked according to $\alpha$, and $k$ is usually a pre-defined pooling ratio (e.g., 0.25, 0.5, 0.75) as it is assumed that only some nodes in each scene-graph are most relevant to the risk assessment task. 
This pooling layer also helps to filter out noise and improve training convergence. 
The node embeddings and edge adjacency information after pooling by $\mathcal X^{pool}_{t}$ and $\mathcal A^{pool}_{t}$ are then calculated as
\begin{equation}
\label{eq:alpha}
\mathcal X^{pool}_{t}= (\mathcal X^{prop}_{t} \odot\mathrm{tanh}(\mathbf{\alpha}))_{\mathcal{P}}, 
\end{equation}
\begin{equation}
{\mathcal A^{pool}_{t}} = {\mathcal A^{prop}_{t}}_{(\mathcal P,\mathcal P)}. \\
\end{equation}
where $\odot$ represents an element-wise multiplication, $()_{\mathcal P}$ indicates the operation of extracting a subset of nodes based on $\mathcal P$, and  $()_{(\mathcal P,\mathcal P)}$ denotes the construction of the adjacency matrix between the nodes in this subset. 
The set of pooled nodes is then processed by a readout layer that compresses the node embeddings $\mathcal{X}^{pool}_{t}$ into a single graph embedding $\mathbf h_{\mathcal G_t}$ as 
\begin{equation}
    \mathbf{h}_{\mathcal G_t} = \textbf{READOUT}(\mathcal{X}^{pool
    }_{t}) 
    %\quad s.t. \quad \textbf{READOUT} \in \{sum, mean, max\}
\end{equation}
where the \textbf{READOUT} operation can be summation (\textit{sum-pooling}), averaging (\textit{mean-pooling}), or selecting the maximum of each feature dimension over the set of node embeddings (\textit{max-pooling}).
% The spatial modeling, pooling, and readout operations are repeated across each graph extracted from each sample in $I$ to produce the sequence of scene-graph embeddings, $\mathbf{h}_G$.

\subsubsection{Temporal Modeling}
We employ a long short-term memory (LSTM)~\cite{hochreiter1997long} network to convert the sequence of scene-graph embeddings to the spatial-temporal embedding $\mathcal Z$. 
For each timestamp $t$, the LSTM updates the hidden state $p_t$ and cell state $c_t$ as 
\begin{equation}
    p_t, c_t = \mathbf{LSTM}(\mathbf{h}_{\mathcal G_t}, c_{t-1}),
\end{equation}
where $\mathbf{h}_{\mathcal G_t}$ is the final \textit{scene-graph} embedding from timestamp $t$.
After the LSTM processes all the scene-graph embeddings, a temporal readout operation is applied to the resulting output sequence to compute the final spatio-temporal embedding $\mathcal Z$ as
\begin{equation}
    \mathcal{Z} = \textbf{TEMPORAL\_READOUT}(p_1, p_2, ..., p_{\mathcal T})
\end{equation}
where the $\textbf{TEMPORAL\_READOUT}$ operation could be the extraction of only the last hidden state $p_{\mathcal T}$ (LSTM-last) or could be a temporal attention layer (LSTM-attn).
Here we integrate an attention mechanism into the LSTM architecture, i.e., LSTM-attn, to boost model performance. 
%\textcolor{red}{Inspired by the work presented in \cite{yu2021scene}, integrating an attention mechanism into the LSTM architecture has been shown to boost its performance. 
By adding an attention layer between successive LSTM layers, the model can weigh the significance of each timestep in the sequence, allowing it to focus more on the most relevant parts of the data.  %This enhanced focus mechanism has proven its utility in various tasks, including the AV (Autonomous Vehicle) risk assessment that we are addressing in this work. 
Additionally, the LSTM-attn layer calculates a context vector by considering the entire hidden state sequence $\{p_1, p_2, ..., p_{\mathcal T}\}$ returned from the LSTM encoder layers, and can hence effectively enhance the model's ability to make accurate and contextually-aware predictions.
%Therefore, 
%Consequently, 
% we incorporated an \textit{LSTM-attn} layer into our model to capture the aggregated information from all timesteps, effectively enhancing the model's ability to make accurate and contextually-aware predictions. 
% This layer is designed to compute a context vector by considering the entire hidden state sequence $\{p_1, p_2, ..., p_{\mathcal T}\}$ that is returned from the LSTM encoder layers.
%The purpose of this context vector is to capture the aggregated information from all timesteps, effectively enhancing the model's ability to make accurate and contextually-aware predictions. 
% \begin{equation}
% \label{eq:temporalattn}
% q = \sum^{T}_{t=1} \beta_t p_t
% \end{equation}
% where the probability $\beta_t$ reflects the importance of $p_t$ in generating $q$.
% The probability $\beta_t$ is computed by a {\it Softmax} output of an energy function vector $e$, whose component $e_t$ is the energy corresponding to $p_t$. Thus, the probability $\beta_t$ is formally given by 
% \begin{equation}
%     \beta_t = \frac{\text{exp}(e_t)}{\sum_{k=1}^T \text{exp}(e_k)},
% \end{equation}
% where the energy $e_t$ associated with $p_t$ is given by $e_t = b(s_0, p_t)$. 
% The temporal attention layer $b$ scores the importance of the hidden state $p_t$ to the model objective (e.g., binary classification for risk assessment).
% The variable $s_0$ in the temporal attention layer $b$ is computed from the last hidden representation $p_T$. The input sequence's final spatio-temporal embedding, $Z$, is produced by feeding the context vector $q$ to an LSTM decoder layer. 

The last layer in our model generates an output risk classification $\hat{\mathcal Y}$ from the spatio-temporal embedding $\mathcal Z$ as 
\begin{equation}
    \hat{\mathcal Y} = Softmax(\mathbf{MLP}(\mathcal Z))
\end{equation}
Since our model is implemented as a binary classifier, we use Cross-Entropy loss to train the model. i.e. %The end-to-end workflow of our approach can be summarized as Algorithm \ref{alg:riskassessment}.
\begin{equation}
    \arg\min \mathbf{CrossEntropyLoss}(\mathcal Y, \mathcal{\hat Y})
\end{equation}

\section{Experimental Results}
\label{sec:rs2g_exp}
In this section, we compare the experimental results of our proposed $\Design$ with SOTA graph-based and DL-based approaches.
We start by presenting our experimental setup, training procedure, and key metrics. 
We then demonstrate results for subjective risk assessment across diverse driving datasets and Sim2Real transfer learning capabilities of each method.
We also provide a detailed ablation study showing the benefits of each component of our $\Design$ modeling. %Finally, we analyze the key differences between the relations and graph structures learned by RS2G and the rule-based relations used by SOTA graph extraction methods.

\subsection{Experimental Setup}
%\subsubsection{Dataset and Platform}
\textbf{Platform:} We conduct our experiments on a Linux server with an Intel Xeon E5 CPU and an NVIDIA TITAN Xp GPU for training and evaluating each model.

\textbf{Dataset:} Our evaluation utilized three different types of datasets: 
(\romannumeral 1) simulated lane change scenarios of varying risk from Carla~\cite{dosovitskiy2017carla}, denoted as \textit{271-carla} and \textit{1043-carla}; 
(\romannumeral 2) real-world, clear-weather safe driving in California Bay Area from Honda~\cite{ramanishka2018toward}, denoted as \textit{1361-honda}; 
and 
(\romannumeral 3) real-world crashes and dangerous road scenarios from dash-cam footage from the Detection of Traffic Anomaly Dataset~\cite{yao2020dota}, denoted as \textit{620-dash}. 
%. We refer to these datasets as (i) \textit{271-carla} and \textit{1043-carla}, (ii) \textit{1361-honda}, and (iii) \textit{620-dash}, respectively, with the number indicating the number of driving clips in each dataset. 
% Each driving clip is between 10-40 seconds in duration. We also used a subset of \textit{1361-honda} consisting of only lane-changing clips, denoted as \textit{571-honda} for our transfer learning experiments.
%For further details about dataset preparation, please refer to \cite{malawade2022spatiotemporal}.
%We use the roadscene2vec library~\cite{malawade2022roadscene2vec} to represent the scene-graphs, train the models, and perform the evaluation in PyTorch. 
We use the same dataset preparation steps as  in ~\cite{malawade2022spatiotemporal}.
The number of risky and non-risky scenes in each dataset is listed in Table \ref{tab:imbalance}.
For subjective risk assessment of each dataset, we use $70\%$ of data for training and $30\%$ of data for inference. 
For transfer learning experiments, we train with 70\% of the training dataset and evaluate with 100\% of the inference dataset, since the training and inference datasets are distinct. 

\begin{table}[!t]
%\small
\footnotesize
\caption{Detailed Breakdowns of Datasets}%\vspace{-0.3cm}
%% \resizebox{1\columnwidth}{!}{
\vspace{-5mm}
\begin{center}
\begin{tabular}{c|c|c|c} \toprule
Dataset 
& \begin{tabular}[c]{@{}c@{}}Non-Risky\\ Scenarios\end{tabular} 
&\begin{tabular}[c]{@{}c@{}}Risky\\ Scenarios\end{tabular} 
& \begin{tabular}[c]{@{}c@{}}Non-Risky:Risky\\ Ratio\end{tabular}  \\
\midrule
\textit{271-carla} & 223 & 48 & 4.65:1\\
\textit{571-honda} &  475 & 99 & 4.80:1\\
\textit{620-dash} &  323 & 297 & 1.09:1\\
\textit{1043-carla} &  898 & 146 & 6.15:1\\
\textit{1361-honda} &  1207 & 154 & 7.84:1\\
%\hline
\bottomrule
\end{tabular}
\label{tab:imbalance}
\end{center}
%% }
\vspace{-3mm}
\end{table}

%\subsubsection{Model Specification}
\textbf{Model Specification:}
Our proposed model consists of three main modules: graph extraction, spatial model, and temporal model.
For graph extraction, we implement three variants of edge encoding methods for $\Design$: %one-dimensional MLP, two-dimensional MLP, and Transformer, each corresponding to a different type of encoder. 
{(\romannumeral 1) one-dimensional MLP, denoted as \textit{RS2G(1D MLP)}, employs a node encoder of dimensions $15\times 15$; given that the edge encoder processes features from two nodes simultaneously, its shape is $30\times 12$. In particular, each node vector is of dimension $15$, and hence the concatenation of two nodes is of dimension 30. We set $12$ as the total number of relations; a trivial number of relations can impede the expressiveness of graph representations, while an overlarge number of relations can weaken graph representations by making the relations redundant.} %if the number of relations is too small, it won't be enough to express the relation graph and if the number of relationships is too large, it will make the relationship diagram redundant, weakening the ability to express the graph.}
(\romannumeral 2) For two-dimensional MLP, denoted as \textit{RS2G(2D MLP)}, both the node and edge encoders have one extra layer, with shapes of $15\times15\times15$ and $30\times30\times12$, respectively. 
(\romannumeral 3) 
For the Transformer variant, denoted as \textit{RS2G(Transformer)}, the node encoder retains the $16\times32\times16$ shape as in the 2D MLP. The Transformer encoder itself is configured with 
% a \textcolor{red}{d\_model}
\texttt{d\_model}
(indicating the expected number of features in the input) set to 32 and the number layer is set to 8. Following this encoder, there's an accompanying MLP with dimensions $32\times12$ to finalize the feature transformation.
%\textcolor{red}{
%For graph extraction, we implement three variants with RS2G: 1D MLP, 2D MLP, and Transformer, each corresponding to a different type of encoder. 
%1D MLP employs a node encoder of dimensions 15x15; given that the edge encoder processes features from two nodes simultaneously, its shape is 30x12. For 2D MLP, both the node and edge encoders have an added depth, with shapes of 15x15x15 and 30x30x12, respectively. 
% For the Transformer variant, the node encoder retains the 16x32x16 shape as in the 2D MLP. The Transformer encoder itself is configured with a d\_model (indicating the expected number of features in the input) set to 32 and nhead (number of self-attention heads) equal to 8. Following this encoder, there's an accompanying MLP with dimensions 32x12 to finalize the feature transformation.}
Our spatial and temporal modeling follow the same structure of MR-GCN and LSTM as the downstream of~\cite{yu2021scene} for fair comparisons.
Specifically, for modeling spatial graph features, we utilize a 2-layer MR-GCN with self-attention graph pooling and \textit{mean} readout. 
We then apply a 2-layer LSTM with temporal attention as the readout operation for our temporal model. 

\textbf{Baseline Models:}
We compare $\Design$ with 
(\romannumeral 1) the SOTA rule-based graph extraction and learning approach~\cite{yu2021scene}, denoted as ``Rule-Based'' graph extraction in our experiments; and 
(\romannumeral 2)
 the SOTA DL-based approach utilizing the CNN+LSTM architecture~\cite{yurtsever2019risky}, which we denote its graph extraction as ``None''
since this method does not use graphs.

%\subsubsection{Evaluation Matrices and Baseline Models}
\textbf{Evaluation Metrics:}
As we model risk assessment as a binary classification task, we evaluate each model in terms of Accuracy, Matthews Correlation Coefficient (MCC) \cite{chicco2020advantages}, and Area Under the ROC Curve (AUC) \cite{bradley1997use}. 
Accuracy in this case is the standard metric indicating the percentage of correctly classified scenes. AUC score is a typical metric for scoring classifiers across multiple decision boundaries; it ranges from $0.0$ to $1.0$ with higher performance indicating a more robust model. MCC score is considered a more reliable metric than accuracy for evaluating models on imbalanced datasets. Specifically, an MCC score of $-1.0$ represents an always wrong classifier, $1.0$ represents an always correct classifier, and $0.0$ represents a random classifier.

% \textbf{Platforms: }
% We used a Linux server with an Intel Xeon E5 CPU and an Nvidia Titan Xp GPU for training and evaluating each model. %Notably, the RS2G models were slower to train since they needed to train the graph encoder layers in addition to the MR-GCN and LSTM layers. Additionally, the edge density of the graphs affected training time since each edge adds graph convolution operations, enabling the sparser rule-based graphs to train faster. Training speed and convergence could be improved in future work if graph sparsity is encouraged during RS2G training (e.g., via additional training objectives).
%\vspace{-1mm}
\subsection{Subjective Risk Assessment}
\label{subsec:risk_assessment}
Table \ref{tab:risk_assess} demonstrates the performance of each model variant at subjective risk assessment on both synthetic datasets (\textit{271-carla, 1043-carla}) and real-world driving datasets (\textit{620-dash, 1361-honda}).
Across all datasets, all the $\Design$ models (1D MLP, 2D MLP, and Transformer) demonstrate significantly higher accuracy than the SOTA DL-based model
 using no graph extractions and the SOTA rule-based graph extraction method, indicating that our data-driven graph extraction technique can effectively enhance graph representations and scenario understanding for AVs. 
Specifically, $\Design$ (Transformer) provides on average $21.35\%$ higher accuracy than the SOTA DL-based approach and $4.47\%$ higher accuracy than the SOTA rule-based graph extraction method. 
Additionally, our data-driven graph extraction method also provides considerably higher MCC and AUC scores, indicating notably better capabilities in distinguishing positive and negative samples. 
Furthermore, $\Design$ (Transformer) notably outperforms other edge encoding methods (1D MLP and 2D MLP) in most cases, proving that the attention mechanism of the Transformer can better capture complex relations among road users.

Overall, all the models provide lower learning quality, i.e., accuracy, MCC, and AUC, for real-world driving datasets than synthetic datasets. 
In particular, for a real-world imbalanced dataset with more crashes and risky scenarios, i.e., \textit{620-dash}, the CNN-based model delivers seriously degraded accuracy and an MCC score worse than a random classifier (less than 0.0). 
However, all the $\Design$ models exhibit notably less performance degradation, indicating that our data-driven graph extraction technique can provide more effective performance in complex real-world scenarios than SOTA DL-based approaches and SOTA GL-based models using rule-based graph extraction method. 
%Additionally, $\Design$ utilizing the edge encoder inspired by the Transformer outperforms other edge encoding methods (1D MLP and 2D MLP) in most cases, proving that the attention mechanism of the Transformer can better capture various relations in a road scene.  
% In particular, the CNN+LSTM with no graph extraction shows a weak MCC score across all datasets (i.e., only slightly better than a random classifier), indicating it cannot distinguish positive and negative instances well. 

%我把这里有关具体模型的说法都改成统称data-drvien method
%Across all datasets, using 1D MLP, 2D MLP and Transformer graph extraction layers provide significantly higher accuracy than those using ``None'' (CNN+LSTM) or rule-based graph extraction. On average, the Transformer extraction technique provides $21.35\%$ higher accuracy than the CNN+LSTM and $4.47\%$ higher accuracy than rule-based graph extraction. Additionally, the data-driven based method also provide higher MCC and AUC scores, indicating that using data-driven graph extraction techniques has a decisive advantage in distinguishing positive and negative samples. In contrast, using none graph extraction with CNN+LSTM provides significantly lower learning quality. In particular, for a real-world imbalanced dataset with more crashes and risky scenarios (\textit{620-dash}), CNN+LSTM delivers seriously degraded accuracy and an MCC score worse than a random classifier (less than 0.0). Data-Driven graph extraction also has reduced performance for \textit{620-dash}, but it is still significantly better than the baselines. 

\begin{table}[ht]
%\vspace{-3mm}
\footnotesize
    \centering
    \begin{tabular}{c c c c  c c c c c}\hline
Dataset & \begin{tabular}[c]{@{}c@{}}Graph\\ Extraction\end{tabular}
& Accuracy & MCC & AUC \\
%\cline{2-3}& Graph Ext. & Downstream &  &  & \\
\hline
\multirow{4}{*}{\textit{271-carla}} & None  & 73.17\% & 0.1887 & 0.8043\\
& Rule-Based  & 82.93\% & 0.5173 & 0.8098\\
& RS2G (1D MLP)  & 84.51\% & 0.2093& 0.9338\\
& RS2G (2D MLP) & \textbf{86.59\%} & \textbf{0.468} & \textbf{0.9578}\\
& RS2G (Transformer) & 
84.15\% & 0.402 & 0.9362\\\hline
\multirow{4}{*}{\textit{1043-carla}}  & None  & 71.66\% & 0.1111 & 0.7173\\
& Rule-Based   & 91.43\% & 0.7217 & 0.971\\
& RS2G (1D MLP) & 91.72\% & 0.6840 & 0.9643\\
& RS2G (2D MLP) & 93.31\% & 0.7426  & 0.7949\\
& RS2G (Transformer) & \textbf{97.13\%} & \textbf{0.8823} & \textbf{0.9686}\\\hline
\multirow{4}{*}{\textit{1361-honda}} & None  & 60.39\% & 0.0391 & 0.7110\\
& Rule-Based   & 86.31\% & 0.2445 & 0.9341\\
& RS2G (1D MLP) & 87.04\% & 0.1626 & 0.9315\\
& RS2G (2D MLP) & 89.00\% & 0.3029 & 0.9383\\
& RS2G (Transformer) & \textbf{89.98\%} & \textbf{0.404} & \textbf{0.9495}\\\hline
\multirow{4}{*}{\textit{620-dash}}& None  & 48.92\% & -0.1749 & 0.5256\\
& Rule-Based   & 67.20\% & 0.3428 & 0.6966\\
& RS2G (1D MLP) & 68.82\% & 0.3967 & 0.7403\\
& RS2G (2D MLP) & \textbf{72.04\%} & \textbf{0.4398} & \textbf{0.8047}\\
& RS2G (Transformer) & 68.28\% & 0.3635 & 0.7354\\
\hline
    \end{tabular}
    %\vspace{-1mm}
    \caption{Performance of Subjective Risk Assessment for different graph extraction methods and datasets. ``Rule-based'' refers to the SOTA GL-based model based on rule-based graph extraction \cite{yu2021scene}, and $\Design$ is our proposed approach.
    The downstream of these methods consists of an MR-GCN and an LSTM model, taking scene-graphs as input.
    ``None'' graph extraction refers to SOTA DL-based model~\cite{yurtsever2019risky} without using graphs, where the downstream task processes raw image data using a CNN and an LSTM model.}
    \label{tab:risk_assess}
   %\vspace{-2mm}
\end{table}

\subsection{Transfer Learning Evaluation}
\label{subsec:transfer}
We evaluate the \textit{Sim2Real} transfer learning capability of each model, i.e., the capability of generalizing knowledge gained from simulations to real-world scenarios. 
Specifically, we first train each model on one of the simulation datasets (\textit{271-carla} or \textit{1043-carla}), and then evaluate the trained model on the real-world dataset (\textit{620-dash}) where there are considerably more instances of crash scenarios. 
Simulations in synthetic datasets include only behaviors of lane changes, while all driving maneuvers are presented in the real-world dataset. In other words, the visual context of the real-world dataset significantly differs from that in simulation environments.
Consequently, as demonstrated in Table \ref{tab:transfer},
% The analysis is two-fold: (i) the driving behaviors differ from the simulation (lane changes only) to the real-world datasets (all driving maneuvers are presented), and (ii) the visual context differs between the simulation environment and the real-world scenarios.  %In particular, we train our models on simulated datasets \textit{271-carla} and \textit{1043-carla} and perform transfer learning on \textit{620-dash}, where there are many instances of crash scenarios. 
%as \textit{620-dash} includes significantly more risky scenarios and differs greatly from the simulated datasets, 
all the learning methods deliver degraded performance.
However, $\Design$ models (2D MLP, Transformer) %exhibit considerably less performance degradation and 
achieve notably higher accuracy, MCC, and AUC than the SOTA DL-based model and the SOTA GL-based method using rule-based graph extraction. 
Specifically, $\Design$ (Transformer) provides on average $14.68\%$ higher accuracy than SOTA DL-based model, $15.90$ higher accuracy than SOTA GL-based method using rule-based graph extraction, and $6.54\%$ than $\Design$ (2D MLP).
% Since 620-dash includes significantly more risky scenarios and differs greatly from the simulated datasets, all learning methods demonstrate degraded performance. However, compared to other methods,  $\Design$ exhibits notably less degradation and delivers considerably higher accuracy, MCC, and AUC. 

It is also noteworthy that the baseline models  
%Notably, the two baselines (CNN+LSTM and Rule-Based MR-GCN) 
trained on \textit{1043-carla} both deliver lower performance after transfer than the same models trained on \textit{271-carla}, which contains notably fewer data samples than \textit{1043-carla}. 
It is likely to be caused by their overfitting to the training domain, i.e., the simulation environments, thus degrading their performance in the test domain, i.e., real-world settings. 
%likely because these models overfit to their training domain (synthetic data) at a detriment to the test domain performance (real-world driving). 
In contrast, our data-driven $\Design$, especially models utilizing the Transformer as the relation extractor, dynamically extract more expressive graph representations and therefore is less likely to encounter overfitting issues, hence providing better performance for Sim2Real transfer learning.

%can extract more expressive graph representation so that generalize well across dataset sizes without overfitting in the same manner, resulting in higher transfer performance when trained on \textit{1043-carla} and \textit{271-carla}. 
\begin{table}[ht]
%\vspace{-1mm}
\footnotesize
    \centering
    \begin{tabular}{c c c c  c c c c c}\hline
Dataset & \begin{tabular}[c]{@{}c@{}}Graph\\ Extraction\end{tabular}
& Accuracy & MCC & AUC 
\\
%\cline{2-3}& Graph Ext. & Downstream &  &  & \\
\hline
\multirow{4}{38pt}{\textit{271-carla to 620-dash}} 
& None & 52.58\% & 0.0333 & 0.5126\\
& Rule-Based  & 48.22\% & 0.0238 & 0.4975\\
& RS2G(2D MLP)  & 57.25\% & 0.1398 & 0.5669\\
& RS2G(Transformer) & 
\textbf{64.68}\% & \textbf{0.2957} & \textbf{0.6831} \\\hline
\multirow{4}{40pt}{\textit{1043-carla to 620-dash}}
& None  & 49.03\% & -0.0432 & 0.4999\\
& Rule-Based  & 50.96\% & 0.0021& 0.5093\\
& RS2G(2D MLP)  & 60.65\% & 0.2089 & 0.6265\\
& RS2G (Transformer) & \textbf{66.29\%} & \textbf{0.3293} & \textbf{0.6964}\\\hline
    \end{tabular}
    %\vspace{-2mm}
    \caption{Transfer learning comparison between different the SOTA DL-based model~\cite{yurtsever2019risky}, the SOTA GL-based model with rule-based graph extraction, $\Design$ (2D MLP) and $\Design$ (Transformer).} %. Results are denoted as ``(Train Dataset) to (Test Dataset)''.}
    \label{tab:transfer}
    %\vspace{-2mm}
\end{table}

\subsection{Ablation Study}
%Here we provide detailed ablation studies to further demonstrate the benefits of Kullback-Leibler(KL) divergence, and results using different edge extraction thresholds. We present the impact of each component of the downstream for GL-based models in the appendix. 

\subsubsection{Analysis of  Kullback-Leibler Divergence}
\label{subsubsec:KL_analysis}
%In Section \ref{subsubsec:graph_gene}, we demonstrate that combining the variational autoencoder (VAE) and KL divergence with the Transformer model enables a more representative scene-graph across diverse driving conditions. 
 We present the effects of KL divergence on $\Design$ (2D MLP) and $\Design$ (Transformer) in Sim2Real transfer learning in Table \ref{tab:kl_analysis}. 
In particular, when the KL divergence is not involved in the model, $\Design$ (Transformer) outperforms $\Design$ (2D MLP) by $1.93\%$. 
However, after incorporating KL divergence into the model, the performance of $\Design$ (2D MLP) slightly drops whereas the performance of $\Design$ (Transformer) noticeably improves. We summarize the potential reason as follows: 
The scene-graphs generated from MLP tend to be simpler %in their representations because of 
due to the simple MLP architecture which limits the model to capture sequential or relational intricacies in data.
Thus, when KL regularization is introduced, it places an extra constraint on a graph that is already simple, pushing the model towards even more conservative graph representations. 
Consequently, the expressiveness of graphs becomes very limited, leading to less effective 
%Consequently, the expressiveness and detailed granularity of the scene-graph is impeded, leading to less effective 
representations of intricate tasks or diverse conditions by $\Design$ (2D MLP).
On the other hand, for $\Design$ (Transformer), given the power of the attention mechanism of the Transformer, 
its generated scene-graphs can be very comprehensive, sometimes at the risk of being overly complex.
Incorporating KL divergence into the Transformer does not simply constrain the model; instead, it refines the Transformer attention, ensuring that the model captures the most representative relation of the adjacency matrix. Thus, combining the Transformer with KL divergence produces more balanced and expressive scene-graphs, and leads to higher accuracy, MCC, and AUC scores. 
%\textcolor{blue}{
%Observing the results, when the KL term is not incorporated, the RS2G(Transformer) outperforms the RS2G(2MLP) by approximately 2\%. 
%However, the integration of KL loss exhibits distinct behavior in these two architectures. For RS2G(2MLP), the performance drops slightly upon incorporating the KL loss, whereas for the RS2G(Transformer), there is a noticeable improvement.
%The scene graphs generated from MLP tend to be simpler in their representations because of the simple MLP architecture which limit itself to capture sequential or relational intricacies in data. Consequently, when KL regularization is introduced, it places an additional constraint on an already simple graph, pushing the model towards even more conservative representations. 
%This can limit the expressiveness and detailed granularity of the scene graph, rendering it less effective for intricate tasks or diverse conditions.
%The Transformer, however, given its attention mechanisms, the scene graphs produced can be quite detailed and intricate, sometimes even running the risk of becoming overly complex. 
%By introducing KL to the Transformer, it doesn't simply constrain the model; instead, it refines its attention, ensuring that the model captures the most representative relation of the adjacency matrix, thereby producing more balanced and expressive scene graphs. 
%This is evident from the performance boost seen in all metrics (Accuracy, MCC, and AUC) for the Transformer with KL, indicating enhanced generalization.}
\begin{table}[ht]
%\vspace{-2mm}
\footnotesize
    \centering
    \begin{tabular}{c c c c  c c c c c}\hline
 \multicolumn{2}{c}{Model} & \multirow{2}{4em}{Accuracy} & \multirow{2}{4em}{MCC} & \multirow{2}{4em}{AUC} \\\cline{1-2}
 Graph Extraction & with KL &  &  & \\\hline
%\multirow{4}{40pt}{\textit{1043-carla to 620-dash}}
 RS2G(2MLP) & \xmark & 62.42\% & 0.2455 & 0.6132\\
 RS2G(2MLP) & \cmark & 60.65\% & 0.2089& 0.6265\\
 RS2G(Transformer) &  \xmark & 64.35\% & 0.2897 & 0.6586\\
 RS2G(Transformer)& \cmark & \textbf{66.29\%} & \textbf{0.3293} & \textbf{0.6964}\\\hline
    \end{tabular}
    \vspace{-2mm}
    \caption{KL Analysis for $\Design$ (2D MLP) and $\Design$ (Transformer) in transfer learning from \textit{1043-carla} to \textit{620-dash}.}
    \label{tab:kl_analysis}
   %\vspace{-1mm}
\end{table}

\subsubsection{Graph Structure Comparison}
%In addition to measuring the similarity of different relation types, 
We compare the structural differences between rule-based graphs and the data-driven graphs extracted by $\Design$. 
In particular, We evaluate how the methods differ regarding graph sparsity and edge distribution, and correlate these metrics with risk assessment accuracy. 
We also identify how  $\Design$'s edge extraction threshold ($\gamma$) affects the sparsity of generated graphs and the model's overall performance. 
Specifically, the threshold $\gamma$ indicates the sigmoid score that $Encode_{edge}$ must overcome to add a given edge to the graph varying from 0 to 1, i.e., higher $\gamma$ results in sparser graphs.
Our results using Transformer graph extraction with different $\gamma$ are shown in Table \ref{tab:structure_comparison}. Transformer extraction with various thresholds exhibits higher accuracy than rule-based graph extraction, and the best performance is achieved with $\gamma = 0.25$. 
%. We achieved the best accuracy using $\gamma = 0.25$, so this setting was used for the rest of the experiments shown in the paper. 
Using $\gamma=0.5$ and $\gamma=0.75$ lowers the performance, possibly due to overfitting. 
On the other hand, using $\gamma=0.75$ can reach a better performance than $\gamma=0.5$ may due to some seeming irrelevant nodes could actually provide useful information for the model. % to make the judgment. 
\begin{table}[h]
\footnotesize
\centering
\begin{tabular}{c c c c c c c c}
\hline
Graph Ext. & Acc. & Avg. Deg. & Avg. Edges & $\sigma$ Edges\\\hline
Rule-Based  & 95.86\% & 3.84 & 16.50 & 10.51\\
RS2G ($\gamma=0.25$) & \textbf{97.13}\% & \textbf{37.11} & \textbf{298.98} & \textbf{264.36} \\
RS2G ($\gamma=0.5$) & 94.59\% & 23.98 & 193.21 & 171.22\\
RS2G ($\gamma=0.75$) & 95.54\% & 10.88 & 87.68 & 78.00 \\
\hline
\end{tabular}
\vspace{-2mm}
\caption{Comparison of graph structure metrics between rule-based graph extraction~\cite{yu2021scene} and $\Design$ (Transformer). $\gamma$ represents the edge extraction decision threshold. }%RS2G(Transformer, $\gamma=0.25$) has the highest accuracy.}
\label{tab:structure_comparison}
\vspace{-5mm}
\end{table}

%实验加alation study解释KL对transfer的影响

%附录里的k的取值要跑Transformer

%2个 20h+ 一天

%cosine表格要transformer的结果
%时间

%实验表格补齐，复测
%3-4h/每个

%figure3改成表格

\section{Conclusion}
In this paper, we propose RS2G, an innovative road scene understanding framework based on data-driven graph extraction and modeling approach, which dynamically captures the diverse relations among road users. 
%In this paper, we propose $\Design$, an innovative data-driven graph extraction and modeling approach that dynamically captures the diverse relations among road users to enhance autonomous scenario understanding. 
In contrast to the rule-based graph extraction method, $\Design$ learns to specialize relations with data-driven vectors, thereby providing more expressive graph representations of road scenes. 
We also leverage the powerful attention mechanism of the Transformer and the variational autoencoder to further enhance $\Design$'s capability to model complex relations and to transfer knowledge from training domains to real-world scenarios. 
Our evaluation demonstrates that $\Design$ significantly outperforms the SOTA DL-based model and the SOTA rule-based graph extraction method in both subjective risk assessment and Sim2Real transfer learning. 

\newpage
%%%%%%%% REFERENCES
{\small
\bibliographystyle{ieee_fullname}
\bibliography{egbib}
}

\end{document}